# Eagle: End-to-end Deep Reinforcement Learning based Autonomous Control of PTZ Cameras

SANDEEP SINGH SANDHA, University of California, Los Angeles, USA

BHARATHAN BALAJI, Amazon, USA

LUIS GARCIA, USC Information Sciences Institute, USA

MANI SRIVASTAVA, University of California, Los Angeles, USA

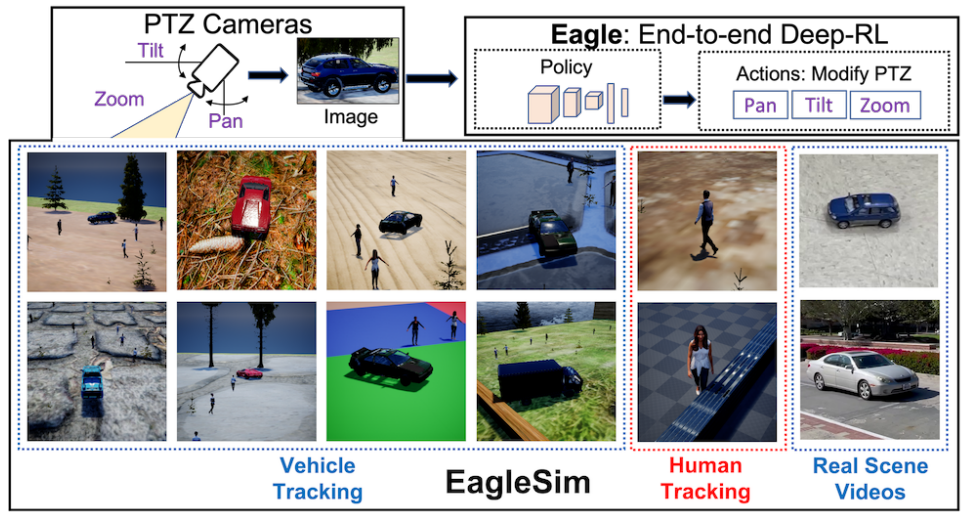

Fig. 1. Eagle trains end-to-end deep-RL controllers for PTZ cameras. Sample scenes for vehicle and human tracking from the EagleSim simulator are shown. The direct transfer of Eagle policies to real-scene videos is also demonstrated.

Existing approaches for autonomous control of pan-tilt-zoom (PTZ) cameras use multiple stages where object detection and localization are performed separately from the control of the PTZ mechanisms. These approaches require manual labels and suffer from performance bottlenecks due to error propagation across the multi-stage flow of information. The large size of object detection neural networks also makes prior solutions infeasible for real-time deployment in resource-constrained devices. We present an end-to-end deep reinforcement learning (RL) solution called *Eagle*[1] to train a neural network policy that directly takes images as input to control the PTZ camera. Training reinforcement learning is cumbersome in the real world due to labeling effort, runtime environment stochasticity, and fragile experimental setups. We introduce a photo-realistic simulation framework for training and evaluation of PTZ camera control policies. Eagle achieves superior camera control performance by maintaining the object of interest close to the center of captured images at high resolution and has up to 17% more tracking duration than the state-of-the-art. Eagle policies are lightweight (90x fewer parameters than Yolo5s) and can run on embedded camera platforms such as Raspberry PI (33 FPS) and Jetson Nano (38

---

[1]Open-source link of Eagle: https://github.com/nesl/Eagle_PTZ_Cameras

FPS), facilitating real-time PTZ tracking for resource-constrained environments. With domain randomization, Eagle policies trained in our simulator can be transferred directly to real-world scenarios[2].

**ACM Reference Format:**
Sandeep Singh Sandha, Bharathan Balaji, Luis Garcia, and Mani Srivastava. 2023. Eagle: End-to-end Deep Reinforcement Learning based Autonomous Control of PTZ Cameras. In *International Conference on Internet-of-Things Design and Implementation (IoTDI '23), May 9–12, 2023, San Antonio, TX, USA.* ACM, New York, NY, USA, 20 pages. https://doi.org/10.1145/3576842.3582366

## 1 INTRODUCTION

Active vision endows applications with the ability to decide 'where to look' at runtime. Autonomous control of pan-tilt-zoom (PTZ) cameras provides superior monitoring with active vision systems by tracking objects of interest in real-time [9, 17]. Active vision systems are increasingly deployed in resource-constrained environments such as remote surveillance [7, 11, 17] and mobile robotics [9, 25].

Autonomous PTZ controllers in prior works have multiple stages, namely detection of objects of interest, tracking their trajectories, and control of PTZ configuration parameters to keep objects in the field-of-view [5, 8, 11, 17]. For example, neural networks are used to identify objects of interest, Kalman filters for predicting trajectories, and a PID controller changes PTZ parameters [6, 15, 25]. The multi-stage pipeline faces the following challenges:

**1. Expensive fine-tuning**: The multiple stages suffer from performance bottlenecks as it is non-trivial to tune each step. For example, tuning the parameters of the Kalman filter requires expert domain knowledge and incurs trial-and-error [14, 16]. Fine-tuning neural object detectors requires bounding box labels [8, 16].

**2. Real-time deployment on resource-constrained platforms**: Even lightweight object detectors (such as YOLO [2, 14]) with several millions of network parameters are too complex for embedded camera platforms. This makes it infeasible to run multi-stage PTZ control algorithms in real-time on platforms [1, 14] with memory and computation constraints.

In contrast to the existing multi-stage pipeline, we propose Eagle, an end-to-end deep reinforcement learning (RL) approach. Eagle trains a neural network policy directly mapping raw images to pan-tilt-zoom actions removing the multiple stages of object detection, localization, and control. Recently, deep-RL has been shown to outperform conventional control for several robotic applications [4, 16]. To the best of our knowledge, we are the first to study an end-to-end perception to control policy for PTZ cameras trained with deep-RL. This is partly due to the challenges of training deep-RL in the real world as it requires large environment interactions, expensive experimental setups, and labeling efforts [4, 16]. To enable successful training of Eagle, we also introduce EagleSim, a simulation framework for placement and control of PTZ cameras in photo-realistic virtual worlds.

In multi-stage pipelines, the adoption of neural network object detectors such as YOLO [2, 14] enables flexible tracking goals and general-purpose object detection in the real world. Although training in simulators is widely studied in deep-RL, transferring simulation policies to the real world is still a challenge [4, 16]. This paper poses and strives to answer the following research questions: (i) *How to design a generalizable end-to-end PTZ controller to track different kinds of objects of interest (e.g., different kinds of vehicles)?*, (ii) *Can we enable flexible tracking goals (e.g., task to either*

---

[2]Videos showing Eagle training and direct transfer to real scenes are available at https://sites.google.com/view/sample-tracking-videos

*track vehicles or humans) at runtime without having multiple control stages?*, (iii) *What is the impact on the PTZ controller performance as the tracking complexity is increased (e.g., complex scenes with multiple objects)?*, and (iv) *What are the limitations of end-to-end PTZ control?*

To answer the above questions, we design generalizable deep-RL controllers and evaluate them on unseen objects/surroundings. EagleSim provides software abstractions to control scene variations with multiple objects (vehicles and human characters) and different surroundings (background materials/patterns and trees) as shown in Figure 1. We demonstrate that these rich scene variations are necessary to bridge the gap between simulation and the real world. We further evaluate the performance of Eagle policies as the tracking complexity is increased. To enable flexible tracking goals, we introduce an extra contextual input along with images during training. Depending on the application's needs, the contextual input modifies the policy behavior at deployments, such as either tracking vehicles or humans. Finally, we show that Eagle policies trained purely in simulation transfer directly to the real videos.

We compare Eagle with three different categories of multi-stage approaches by adopting state-of-the-art choices in object detection, tracking, and control stages. More specifically, we evaluate: (i) *Object_detection+tracking (Kalman filter)+control*, (ii) *Object_detection+reinforcement learning*, and (iii) *Relative_location+control*. Our results show that Eagle outperforms current approaches across a suite of vehicle tracking scenarios to maintain superior PTZ control performance and achieve up to 17% more tracking duration. Eagle trains lightweight neural network policies (79k model parameters and 320KB model size) that are real-time deployable on resource-constrained embedded cameras [1] having computation limitations of Raspberry PI 4B [20] (33 FPS) and Jetson Nano [19] (38 FPS) class devices.

In summary, we make the following contributions:
- We present Eagle, an end-to-end deep-RL approach to control a PTZ camera directly using raw images. Eagle is lightweight and doesn't require multi-stage fine-tuning.
- We introduce a new simulator, called EagleSim, to study PTZ cameras in photo-realistic virtual worlds. We also show the direct transfer of Eagle policies trained purely in EagleSim simulator to the real scene videos.
- We compare Eagle with recent works showing its superior performance and evaluating its generalizability across different object/surrounding variations.

## 2 BACKGROUND AND RELATED WORKS

### 2.1 Autonomous Control of PTZ Cameras

Our objective is to keep one or more objects of interest in the field-of-view (FoV) of the PTZ camera at high resolution. An autonomous controller achieves this by modifying the pan/tilt, keeping the desired object in the center of the captured image, and zooming without clipping the object. Keeping an object in the center of the image avoids target loss during sudden movement/direction changes.

Figure 2 represents the different steps in autonomous PTZ control for four classes of approaches using a sample object of interest (car). We compare the recently proposed approaches that use learning-based components for active tracking. Luo et al. [16] have shown that neural network-based active trackers outperform traditional trackers like MIL, Meanshift, and KCF. Each approach is represented from the input image ❶ to the final control ❾.

**Object_detection+tracking+control [5, 15, 25]**: Steps ❶❷❸❹❾. The objects of interest are detected in the image, followed by a short-term tracking algorithm to predict their location in the future frame. The controller adjusts the pan-tilt-zoom to track the objects of interest. Bernardin et al. [5] focus on human targets. They use a face detector and a mean-shift tracker in combination with expert knowledge to control a PTZ camera. Unlu et al. [25] present the control of a PTZ camera for UAV tracking where a ResNet-based object detector is used. Lopez et al. [15] track less frequent objects using a PTZ camera with a faster R-CNN object detector. The object occurrence probability identifies the less frequent objects of interest, and a rule-based controller modifies the PTZ parameters. Using neural network object detectors makes it infeasible to deploy this pipeline on embedded camera platforms [14]. Further, multi-stage information flow necessitates fine-tuning of each stage, and its performance is impacted by errors in each stage [14, 16]. E.g., the object detector's errors impact the tracking performance [14].

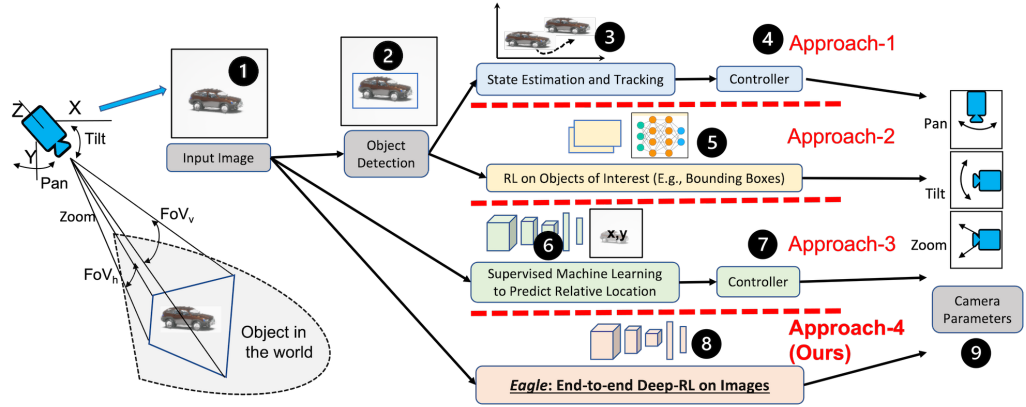

Fig. 2. Different approaches to control PTZ cameras are illustrated using a vehicle tracking scenario. A PTZ camera is controlled to keep a car in the field-of-view (FoV). The horizontal FoV ($FoV_h$) and vertical FoV ($FoV_v$) control zoom parameters. Approach-1 (*Object-detection+tracking+control*): Represented by 1,2,3,4,9 is the widely used multi-stage technique of identifying objects (using object detectors), followed by a short-term tracker and a controller. Approach-2 *Object-detection+RL*: Given by 1,2,5,9 shows a setting where the bounding boxes are used to train an RL policy. Approach-3 *Relative-location+control*: Steps 1,6,7,9 shows an alternative to bounding boxes where a neural network predicts the relative location of objects that the controller uses. Approach-4 *Eagle: End-to-end deep-RL*: Steps 1,8,9 show the proposed Eagle approach to directly control the pan, tilt, and zoom parameters using raw images.

**Object_detection+reinforcement learning [7, 13]**: Steps ❶❷❺❾. These methods combine the tracking and control stages with a deep-RL policy. Bisagno et al. [7] and Kim et al. [13] show the control of PTZ camera using deep-RL where the inputs to the neural networks are the information about the object of interest (e.g., bounding-boxes, location of pedestrians). An actual deployment may need to use external object detectors to measure these inputs, where the performance suffers from object detector errors.

**Relative_location+control [12, 14]**: Steps ❶❻❼❾. A neural network is trained using supervised machine learning to output the relative location of the object of interest in captured images. The relative location is used to control the pan-tilt-zoom parameters without an explicit object detector. However, training the relative location predictor from images demands the availability of labeled bounding boxes in videos which are difficult to get for arbitrary deployments. Further, a separate controller that uses the relative locations needs to be fine-tuned for the specific scenario and camera parameters to make an end-to-end system.

**Eagle**: Steps ❶❽❾. We propose Eagle, an end-to-end deep-RL approach that directly uses the raw input images to control the PTZ parameters of a camera. Luo et al. [16] propose end-to-end deep-RL for first-person tracking, where the first-person observer moves along the object to track. In contrast, we study end-to-end deep-RL to control PTZ cameras where the location of the camera is fixed, but its PTZ parameters are modified to keep an object of interest in the field-of-view. Eagle removes the need to tune the multiple stages and trains lightweight controllers having superior PTZ tracking performance.

## 2.2 Frameworks for Pan-Tilt-Zoom Cameras

Due to the difficulty of creating tracking scenarios in the real world, researchers have proposed several PTZ frameworks. Chen et al. [8] propose a framework where a virtual PTZ camera is controlled to generate images from panoramic videos. However, this framework depends on the human annotation of videos for ground truth. Further, creating new scenarios requires manual video capture using specialized spherical cameras. Hanoun et al. [11] propose a framework to study PTZ camera placement using a CAD environment. However, they assume that the objects of interest are available (similar to the object detectors). Salvagnini et al. [21] propose a framework by placing a real PTZ camera and a calibrated projector screen. This requires specialized equipment, and the spherical screen limits the camera motion. Hamesse et al. [10] propose a PTZ tracker for air traffic control. The simulator doesn't keep track of ground truth annotation; instead, it uses external object detectors and achieves only 3 FPS on a GPU server used by authors. Further, the proposed simulator lacks rich scene variations. It is also unclear how to use existing simulation frameworks for end-to-end deep-RL due to a lack of capability to learn from trial and error in the scenes and automatic ground truth annotations for reward calculations.

**EagleSim**: We introduce the EagleSim simulator to train end-to-end deep-RL policies for PTZ control. EagleSim relieves the need to create real-world tracking scenarios and automatically provides ground truth annotations of objects of interest. These ground truth annotations are perfect, unlike human labels or pre-trained object detectors. We address the challenging simulation-to-reality gap by including virtual words with a wide variety of scene variations. EagleSim includes the capability to simulate pan, tilt, and zoom on real videos and enables parallel scenes supporting >200 FPS on a GPU machine (GeForce RTX 3090 Ti).

## 3 EAGLE: END-TO-END DEEP-RL FOR PTZ

Eagle trains lightweight neural network policies that map input images directly to pan, tilt, and zoom actions to track an object of interest in a scene. We consider a standard Markov decision process where an agent learns from trial and error by interacting with an environment over many discrete time steps. At each step, the agent receives a scalar reward defining its performance. The agent uses the current state of the environment to decide the action.

### 3.1 State Space, Policy Network and Actions

The current state is represented by the most recent image captured from a PTZ camera. The Agent uses the image to decide the next action, where the video frame rate determines the step length. Our policy adopts lightweight network architecture to enable real-time inference on resource-constrained devices. We use a discrete action space that modifies the current values of pan, tilt, and zoom parameters. The implementation-specific details of the input size, state transition delays, network architecture, and action space are discussed in Section 5.2.

### 3.2 Reward Function: Single Object

First, we formalize the reward function to track a single object of interest in scenes having no other objects. This setting is later generalized to handle the presence of other objects.

For active tracking, we formulate a reward function to keep object $O^a$ near the center of the image with the maximum possible resolution. Consider a scenario in Figure 2, where a PTZ camera tracks a car as the object of interest $O^a$. A sample image of $O^a$ captured by the PTZ camera is shown in Figure 3 along with the bounding box of the car $[Xmin, Ymin, Xmax, Ymax]$, image $Height$, and image $Width$. The center coordinate (x,y) of the bounding box is defined as $[x, y] = [(Xmin + Xmax)/2, (Ymin + Ymax)/2]$.

Knowing the bounding box and its center coordinate for $O^a$, we define the reward ($r_i^a$) for step $i$ in Equation 1.

$$r_i^a = \begin{cases} Center_x^a \times Center_y^a \times Obj_{size}^a \times Clip^a - P & Condition \\ -L & Otherwise \end{cases} \quad (1)$$

Where $Condition$ is a binary value. For a single object setting, the $Condition$ is given by $l^a$. Here, $l^a = 1$, if $O^a$ is captured in the image, else $l^a = 0$. When $Condition$ is True, the reward is a multiplication of four terms $Center_x^a$, $Center_y^a$, $Obj_{size}^a$, and $Clip^a$ after penalty $P$ is subtracted. $P > 0$ is a hyper-parameter that penalizes the agent for modifying the camera's PTZ parameters to avoid jittery behavior. When the object $O^a$ is not present in the captured images, the agent receives a negative reward $L$, where $L > 0$ is a hyper-parameter.

$$Center_x^a = \frac{abs\left(\frac{Width}{2} - x\right)}{\frac{Width}{2}} \quad \Big| \quad Center_y^a = \frac{abs\left(\frac{Height}{2} - y\right)}{\frac{Height}{2}} \quad \Big| \quad Obj_{size}^a = \frac{(Xmax - Xmin) \times (Ymax - Ymin)}{Width \times Height} \quad (2)$$

$$Clip^a = \begin{cases} M & \text{if } Xmin = -(Height/2) \text{ or } Ymin = -(Width/2) \\ M & \text{if } Xmax = (Height/2) \text{ or } Ymax = (Width/2) \\ 1 & \text{otherwise} \end{cases} \quad (3)$$

$$R^a = \sum_{i=0}^{N} (r_i^a) \quad (4)$$

$Center_x^a$, $Center_y^a$ and $Obj_{size}^a$ are defined in Equation 2. $Center_x^a$ and $Center_y$ measure the accuracy with which an agent keeps object $O^a$ centered on the X-axis and Y-axis of the image, respectively. If the target is not close to the center of the image, its probability of leaving FoV is high on sudden direction changes. $Obj_{size}^a$ measures the relative size of $O^a$ in the image. $Center_x^a, Center_y^a, Obj_{size}^a \in [0, 1]$. $Clip^a$ is defined in Equation 3, where $M \in (0, 1)$ is a hyper-parameter. $Clip^a$ penalizes the reward $r_i^a$ when $O^a$ is clipped in the captured image. The total reward $R^a$ for each episode of $N$ steps is the summation of step rewards $r_i^a$ as shown in Equation 4.

### 3.3 Generalizable PTZ Tracking

Next, we answer the question: *How to design a generalizable end-to-end PTZ controller to track different kinds of objects of interest?* We extend the reward function to complex scenes by modifying the *Condition* definition in Equation 1.

We define 2 classes of objects in a scene. The first class $A = \{O^{a1}, ..., O^{aT}\}$ is a collection of objects to track. For example, class $A = \{SUV_{blue}, SUV_{red}, Sports_{red}, Sports_{grey}, Pickup_{red}, Pickup_{grey}\}$ is representing different vehicles to track. The second class $B = \{O^{b1}, ..., O^{bU}\}$ is the collection of objects to be ignored. For example, the objects in the class $B$ may refer to background buildings, trees, and human characters for a vehicle tracking scenario. We assume that only one of the objects from class $A$ is present at a given time in the scene, while the same policy generalizes to all objects of class $A$. For example, the same policy can track $SUV_{blue}$, $Sports_{red}$ or a $Pickup_{grey}$ vehicle, but only one of them is present in the scene. The agent is given a reward only when the objects of class $A$ are tracked. We express this by modifying the *Condition*, which is True when $(l^a = 1) \wedge (a \in A)$.

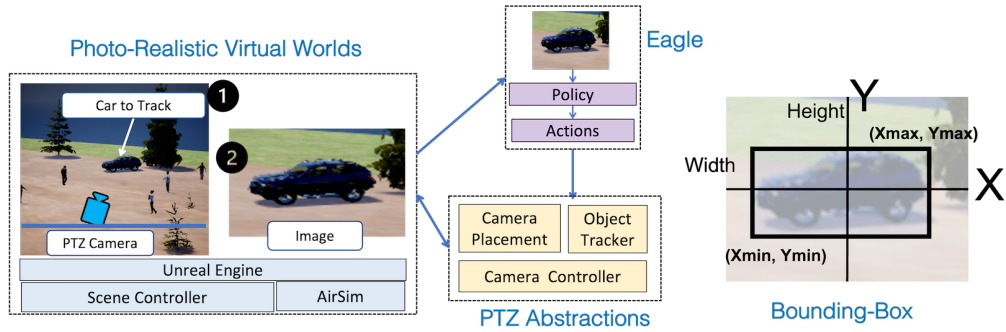

Fig. 3. The architecture of EagleSim and its integration with Eagle along with bounding-box for reward calculation. Step-1 shows a placement of a PTZ camera for vehicle tracking. Step-2 shows an image captured by a camera.

We update the calculation of reward ($r_i^a$) (Equation 1) using this new *Condition*. Our approach to generalize PTZ tracking is motivated by domain randomization (or environment augmentations) [4, 16]. The discussion also applies to other scenarios, such as tracking human characters. Here, the variations of human characters are added to class A, and class B contains objects to be ignored in the scene. When multiple objects of class A are present in the same scene, a policy trained using Equation 1 tracks an object giving better future rewards as discussed in Section 6. Next, we introduce the idea of dynamic tasking to enable flexible tracking Goals.

### 3.4 Dynamic Tasking: Flexible Tracking Goals

Here, we present a novel approach to enable flexible tracking goals at runtime without having multiple control stages. We define dynamic tasking as the capability to change the tracking goal at deployment. Dynamic tasking in multi-stage approaches can be easily realized by changing the object detection stage. The identified objects from the object detectors are filtered during deployment to change the tracking behavior [17]. However, in Eagle, there is no separate object detector. To allow dynamic tasking, we include an extra *contextual input* along with the current camera image and modify the reward calculation.

We use structured contextual input in the integer space to specify different sub-class of objects in class A. To simplify notation, we explain dynamic tasking with an example of two sub-classes in class $A = \{SUV_{blue}, SUV_{red}, Human_1, Human_2\}$. The first sub-class of vehicles $A_v = \{SUV_{blue}, SUV_{red}\}$ and the second sub-class of human characters $A_h = \{Human_1, Human_2\}$. The contextual input $CI \in \{0, 1\}$ represents two integer values. The formulation generalizes to more complex contextual inputs at the expense of increased training time. Our goal is to task policy at runtime to either track sub-class $A_v$ when $CI = 0$ or sub-class $A_h$ when $CI = 1$. We define a new *Condition* in the Equation 5 to allow this.

$$Condition = \begin{cases} True & C_v \vee C_h \\ False & \text{otherwise} \end{cases} \quad (5)$$

Where $C_v$ is True when $(l^a = 1) \wedge (a \in A_v) \wedge (CI = 0)$ and $C_h$ is True when $(l^a = 1) \wedge (a \in A_h) \wedge (CI = 1)$. The *Condition* in the Equation 5 brings in the domain knowledge to ensure that the agent can learn to associate a specific contextual input with a particular sub-class of objects of interest.

## 4 DESIGN OF EAGLESIM

### 4.1 Approaches to training Eagle

Different alternative approaches applicable to train Eagle include:

(i) Training using specialized videos: It is not possible to use videos captured from a standard camera. Training requires videos from different perspectives as modifying pan-tilt-zoom changes the camera's field-of-view. Hence, it either requires multiple cameras to create a 3D world model or specialized spherical cameras to capture panoramic videos [8].

(ii) Training in the real world: Deep-Rl training using a real-world PTZ camera setup is difficult due to a large number of environment interactions, real-time labeling, and fragile experimental setups. Further creating tracking scenarios with vehicles, and humans are not trivial in the real world.

(iii) Training in Simualtors: Researchers [4, 16] have proposed the design of simulation environments to scale the training of deep-RL. However, creating simulations with rich scenarios and transferring policies from simulation to the real world is itself non-trivial.

For optimal real-world performance, a hybrid approach can also be used, where the initial policy is trained using a simulator, and then the real world is used to finetune the policy. We follow approach (iii) and introduce a new simulator called EagleSim to train end-to-end deep-RL policies for PTZ cameras. EagleSim is designed using Unreal Engine[3] and AirSim [24] to enable PTZ camera placement and control in photo-realistic virtual worlds. The architecture of EagleSim is shown in Figure 3.

### 4.2 Photo-Realistic Virtual Worlds

Although AirSim enables control of a drone and a car in virtual worlds, it doesn't support different types of vehicles, control over human characters, background trees, boundaries, and surroundings. These scene variations are needed to create a rich tracking scenario. Creating rich scene variations is non-trivial, and requires a significant engineering effort. To address this, we design new virtual worlds with PTZ software abstractions (4300 lines of c++ code in Unreal engine) and include them with EagleSim.

**Virtual worlds in EagleSim**: We package virtual worlds for vehicle tracking and human tracking scenarios. The vehicle tracking virtual world keeps track of ground truth annotations of vehicles, whereas the human tracking world keeps track of human characters. Each virtual world supports 6 types of vehicles (SUV1, Pickup, Sports, HatchBack, SUV2, Truck) in five colors (30 different vehicles), 6 human characters, 25 background materials/patterns, and 10 types of trees. EagleSim also supports image augmentations to add random shadows, salt-pepper noises, random contrast, and brightness changes to the PTZ images.

Each virtual world represents an open space (70 meters × 70 meters) with boundaries. The boundaries can be made invisible, creating a simple scene with a blue skyline (see Sc-1 setting in Figure 4a). The background patterns can be applied to the floor and the boundaries to create variations of urban, forest, and rural areas, as shown in Figure 1 and Figure 4a.

**Addressing simulation-to-reality gap**: A significant challenge with policies trained in a simulator is to transfer them to the real world. The research community represents this as the simulation-to-reality gap [4, 16] due to sensing and environmental differences between the simulators and the real world. For example, the images in the real world can have observations (such as object variations, backgrounds, shadows, occlusions, and illuminations) never observed in the static/simple simulations. The virtual worlds packaged with EagleSim are carefully designed to address the simulation-to-reality gap. We use domain randomization to train a policy that can generalize to unseen variations of objects and surroundings. The idea is to train policies on a combination of different scene variations.

---

[3]https://www.unrealengine.com/

| Layer | 1 | 2 | 3 | 4 | 5 | 6 | 7 | 8 |
|---|---|---|---|---|---|---|---|---|
| Config | C5x5-64 | C3x3-32 | C3x3-32 | C3x3-16 | 64 | 64 | 64 | 3,3,3 |

Table 1. Policy architecture used by Eagle.

### 4.3 PTZ Abstractions

The PTZ abstractions in EagleSim provide four modular components as shown in Figure 3. A PTZ camera (step ❶) is placed using *Camera Placement* component. The camera is controlled using the *Camera Controller*. Step ❷ shows a sample camera image. The car may or may not get captured in the image depending on the camera's location, the car's location, and PTZ parameters. The bounding box (❸) is captured by the *Object Tracker*.

*Camera Controller* exposes control of pan, tilt, and zoom parameters of the cameras at runtime. The resolution supported for pan and tilt is one degree. The zoom is controlled by modifying the horizontal field-of-view (FoV) at runtime with a resolution of 1 degree. The vertical FoV is based on the aspect ratio and the horizontal FoV: $vertical_{FoV} = horizontal_{FoV} * height/width$. *Object Tracker* provides bounding boxes of objects of interest captured in the PTZ image. A bounding box is calculated by transforming the 3-D outer coordinate mesh of the object available from the Unreal engine to the 2-D image coordinates using a pinhole camera model. The bounding boxes of EagleSim are perfect, unlike the bounding boxes predicted by a neural network-based object detector, which may have prediction errors. The bounding boxes are used to calculate rewards (Section 3) to train Eagle. *Scene Controller* selects scene variations to enable the evolution of the training environment.

## 5 EVALUATION

While every component of EagleSim is needed for end-to-end RL, traditional ablation study is not applicable. Here, we systematically evaluate Eagle policies in varying scene complexity in the virtual world and compare them to the existing multi-stage approaches.

### 5.1 Performance Metrics for PTZ Tracking

Eagle doesn't produce intermediate bounding boxes, generally available in multi-stage approaches. Hence, we adopt metrics (% Tracking, $Center_x$, $Center_y$, $Obj_{size}$) to directly evaluate the camera control performance. The metric of tracking duration (referred to as % Tracking) [14] measures the duration for which the controller successfully keeps an object of interest in the FOV of the camera. When % Tracking is 100, the object was kept in the camera's FoV for the entire duration. Tracking duration is also equivalent to the episode length adopted by Luo et al. [16]. Instead of the number of steps, we report the percentage. Like the tracking duration, Chen et al. [8] uses *track fragmentation*, which is the number of steps as a fraction between 0 and 1. One of the goals of the PTZ controller is to capture an object in the center of the image. To measure center location error, we directly use the average $Center_x$ (Equation 2) and average $Center_y$ (Equation 2) maintained by the controller for the entire trajectory. The metrics of $Center_x$ and $Center_y$ are equivalent to the center location error used by Chen et al. [8]. To compare the object resolution, we adopt $Obj_{size}$ (Equation 2), which measures the relative size of an object in the captured image.

| Scenario | Tracking Goal | Scene Variations |
|---|---|---|
| *Sc-1* | $SUV1_{blue}$ | Fixed background |
| *Sc-2* | $SUV1_{blue}$ | Fixed background+Trees+Image augmentations |
| *Sc-3* | $SUV1_{blue}$ | Variable backgrounds+Trees+Image augmentations |
| *Sc-4* | $SUV1_{blue} + SUV1_{red} + SUV1_{grey}$ | Variable backgrounds+Trees+Image augmentations+Humans |
| *Sc-5* | $SUV1_{blue} + SUV1_{red} + Pickup_{grey} + Pickup_{red} + Sports_{blue} + Sports_{grey}$ | Variable backgrounds+Trees+Image augmentations+Humans |
| *Dynamic Tasking* | $SUV1_{blue}/Humans$ | Variable backgrounds+Trees+Image augmentations |

Table 2. Different tracking scenarios in the increasing order of tracking complexity to evaluate Eagle. The goal of Sc-1 to Sc-5 is to track vehicles. Dynamic tasking (DT) trains a policy to track either a vehicle or human characters.

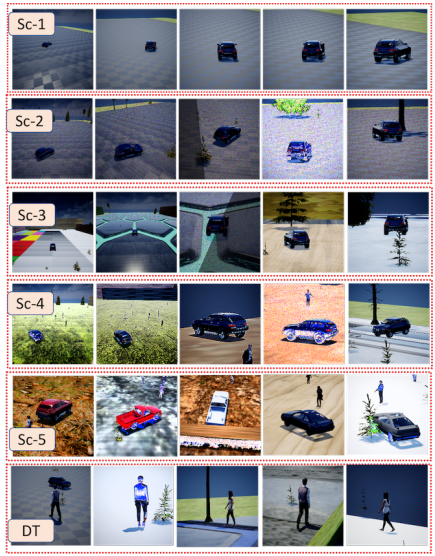
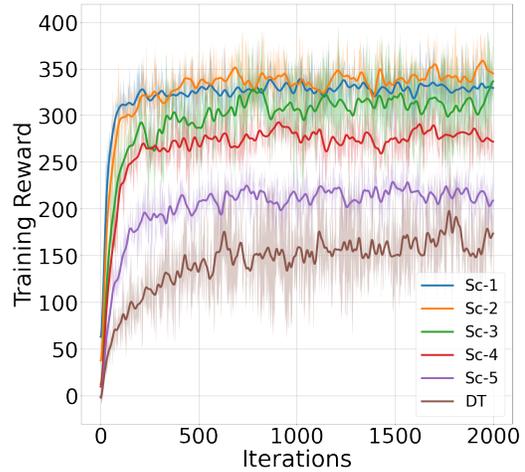

(a)                                                                                                                (b)

Fig. 4. (a) Visualization of tracking scenarios presented in Table 2. Sc-1 to Sc-5 are vehicle tracking scenarios. DT shows the scenes for dynamic tasking of policy to track human characters. (b) Average training reward of Eagle policies for scenarios. We calculate the average reward by training three policies for each scenario and show its min-max spread.

## 5.2 Implementation of Eagle

**State Space, Policy Network and Actions**: We downsample the PTZ camera images to 120×120 and convert them to grayscale. Working directly with large color images is computationally demanding [14, 18]. The low-resolution images are not for surveillance and are only used for the purposes of camera control. Our evaluation in Section 5.4 shows that even at low resolution Eagle outperforms other approaches using higher image resolutions. The policy network architecture is shown in Table 1, which consists of 7 hidden layers and an output layer. This network architecture is motivated by neural networks trained for Atari games by Mnih et al. [18]. The first five layers are convolution layers, each with a stride 2 and a rectifier nonlinearity. The first layer, denoted by C5x5-64 convolves 64 filters of 5 × 5 each. The fifth, sixth, and seventh layers are fully connected and consist of 64 rectifier units each. The last layer produces 3 discrete outputs to control the pan, tilt, and zoom parameters. The policy network is lightweight, enabling real-time inference on embedded platforms (e.g., Raspberry PI 4B, Jetson Nano), and has 79k model parameters.

|  | Fixed background | | | | Variable backgrounds+Trees | | | | Variable backgrounds+Trees+Humans | | | |
|---|---|---|---|---|---|---|---|---|---|---|---|---|
|  | %Tracking | $Center_x$ | $Center_y$ | $Obj_{size}$ | %Tracking | $Center_x$ | $Center_y$ | $Obj_{size}$ | %Tracking | $Center_x$ | $Center_y$ | $Obj_{size}$ |
| Sc-1 | 99.6 ± 4.3 | 0.87 | 0.85 | 0.31 | 9.7 ± 6.3 | 0.65 | 0.63 | 0.13 | 8.5 ± 4.4 | 0.64 | 0.62 | 0.08 |
| Sc-2 | 99.9 ± 0.1 | 0.87 | 0.86 | 0.30 | 18.8 ± 16.2 | 0.81 | 0.79 | 0.29 | 17.2 ± 14.8 | 0.81 | 0.77 | 0.27 |
| Sc-3 | 99.1 ± 8.5 | 0.87 | 0.86 | 0.30 | 98.9 ± 6.9 | 0.87 | 0.86 | 0.30 | 90.0 ± 22.1 | 0.86 | 0.85 | 0.31 |
| Sc-4 | 99.7 ± 4.3 | 0.87 | 0.86 | 0.29 | 98.9 ± 7.4 | 0.86 | 0.86 | 0.29 | 99.1 ± 6.3 | 0.86 | 0.86 | 0.29 |
| Sc-5 | 99.7 ± 5.4 | 0.86 | 0.85 | 0.27 | 99.3 ± 6.9 | 0.86 | 0.86 | 0.27 | 99.2 ± 4.6 | 0.86 | 0.85 | 0.27 |

Table 3. The performance of Eagle policies for vehicle tracking scenarios. $Center_x$, $Center_y$ and $Obj_{size}$ have std within ±0.2.

|  | $HatchBack_{green}$ Vehicle | | | | $Truck_{blue}$ Vehicle | | | |
|---|---|---|---|---|---|---|---|---|
|  | %Tracking | $Center_x$ | $Center_y$ | $Obj_{size}$ | %Tracking | $Center_x$ | $Center_y$ | $Obj_{size}$ |
| Sc-4 | 83.0 ± 28.2 | 0.85 ± 0.1 | 0.86 ± 0.1 | 0.21 ± 0.1 | 92.8 ± 19.4 | 0.87 ± 0.1 | 0.82 ± 0.1 | 0.27 ± 0.1 |
| Sc-5 | 88.1 ± 25.4 | 0.84 ± 0.1 | 0.85 ± 0.1 | 0.22 ± 0.1 | 95.1 ± 16.3 | 0.84 ± 0.1 | 0.83 ± 0.1 | 0.34 ± 0.1 |

Table 4. Generalization of Eagle policies to unseen variations in the object of interest.

The 3 outputs modify the current parameter values as follows: [-2 degrees, 0 degrees, 2 degrees] for pan and tilt. We control the FoV with three possible outputs for the zoom configuration: [-1 degree, 0 degrees, 1 degree]. The horizontal FoV and vertical FoV are equal due to the same aspect ratio of input images. We use the ground truth bounding boxes from EagleSim to calculate the episodic reward (Equation 4) and use $L = 10$, $M = 0.3$, and $Penalty = 0.01$ as the hyperparameters.

**State Transitions**: During training, we maintain an end-to-end delay from sensing a state (image) to action at 30 milliseconds. This delay is matched to the embedded PTZ camera platform [1] when deploying Eagle policy using Raspberry PI 4B. We use an episode length of 2000 steps during training which translates to one minute of continuous tracking.

**Distributed Training**: EagleSim supports the creation of multiple scenes in parallel using Python wrappers in the OpenAI Gym[4] format. This allows the integration of EagleSim with state-of-the-art reinforcement learning libraries. We use 6 parallel scenes to train a single policy in 69 hours (2.9 days) on a GPU server machine (GeForce RTX 3090 Ti) [3] (see Figure 4b). Without 6 parallel scenes, it would take 17 days (2.9*6) for an agent to collect the same amount of data from a sequential environment.

**Training Algorithm**: We adopt distributed implementation of the Proximal Policy Optimization (PPO) [23] algorithm from the stable-baselines3[5] library. The default hyperparameters of PPO present in the library are used, except the updates to the network are performed every 24576 (4096*6) steps collected from 6 parallel environments and a batch size of 256.

### 5.3 Tracking Scenarios

We consider five vehicle tracking scenarios (Sc-1 to Sc-5) with increasing tracking complexity and a dynamic tasking (DT) scenario, as shown in Table 2. The scenes from tracking scenarios are shown in Figure 4a. Sc-1 tracks a single $SUV1_{blue}$ car in a *Fixed background*. In Sc-5, the agent tracks any of the 6 vehicles ($SUV1_{blue}$, $SUV1_{red}$, $Pickup_{grey}$, $Pickup_{red}$, $Sports_{blue}$, $Sports_{grey}$) in the presence of *variable backgrounds*, *tree*, *image augmentations* and *human characters*. *Variable backgrounds* refers to the random selection of background materials (any one of 25 materials included in EagleSim) during training episodes. Various trees and human characters are randomly placed in the scene when enabled. Dynamic tasking (DT) trains a policy to track objects from one of the two sub-classes based on contextual input. The first sub-class contains a single vehicle ($SUV1_{blue}$), and the second sub-class contains 4 different human characters.

---
[4]https://github.com/openai/gym
[5]https://github.com/DLR-RM/stable-baselines3

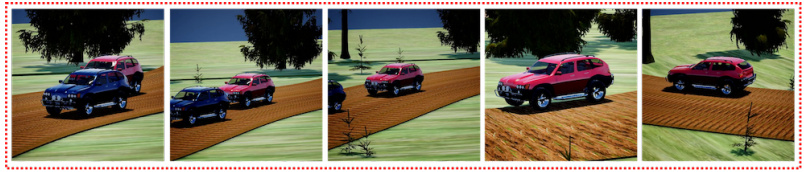

Fig. 5. Policy behavior in a sample scene where the object of interest is not dominant in the initial view and is partially blocked. The goal is to track the red car.

**Tracking setup**: During training and evaluation, the vehicles are given random trajectories by controlling steering and throttle in the virtual world in an open space of 70 meters × 70 meters. A vehicle has an average speed of 6 m/s. The vehicle comes to a standstill when reaching the boundary and randomly changes its direction, and has a max speed of 16 m/s. A PTZ camera is placed at the mid-point of the south (or bottom) boundary at the height of 8 meters as shown in Figure 3 (step ❶). The camera initially looks at the wider scene, capturing a zoomed-out image with a vehicle in it. As the vehicle moves, the goal is to track and focus on it. This tracking setup is shown for different scenes in Figure 4a (Sc-1 to Sc-5), where the initial image shows a wider view of the scene, and subsequently, the PTZ camera tracks the car. With the tracking progress, the PTZ camera focuses on the object of interest, as seen in the right images of Figure 4a. While tracking the vehicle, variations are enabled in specific scenarios, as seen in Figure 4a. For the dynamic tasking (DT) scenario, a human character is placed in the scene along with the vehicle ($SUV1_{blue}$), where both are initially visible to the wider view of the camera. Both humans and vehicles move on random trajectories. A random human character (out of 4) is selected during each training episode.

**Training reward**: The training reward is shown in Figure 4b. The policies are trained using 6 parallel environments for 69 hours (2000 iterations). The training reward is calculated as an average of three policies for each scenario. As seen, the simpler tracking scenario converges faster. We also see with the increasing complexity of scenarios, Eagle policies converge to a lower reward. Next, we analyze the performance of policies to understand this behavior.

*5.3.1 Performance of Eagle Policies.* We evaluate the policies using the checkpoint with the highest training reward. For each scenario, three policies with different random seeds are trained, and a checkpoint is evaluated from each policy. Each checkpoint is evaluated for 100 episodes (each episode is of 2000 steps or 1 minute of tracking). We report the mean performance metrics (Tracking duration (% Tracking), $Center_x$, $Center_y$, $Obj_{size}$) and their std.

Table 3 shows the performance of Eagle policies to track $SUV1_{blue}$ in scenes having *Fixed background*, *Variable backgrounds+Trees*, and *Variable backgrounds+Trees+Humans*. The $SUV1_{blue}$ vehicle is present during training of all scenarios (Table 2). The test scenes are different from training due to the random placement of objects (backgrounds, trees, and human characters when enabled) and random vehicle/human trajectories.

**Policy behavior in simple scenes and understanding training rewards**: Looking at the *Fixed background* evaluation in Table 3, we see Eagle policies can successfully track >99% of the time and maintain similar $Center_x$ and $Center_y$ metrics. However, the policy zooms in conservatively on the object of interest when the training complexity of scenes is increased. The $Obj_{size}$ maintained by policies decreases for Sc-1 to Sc-5 gradually. Our hypothesis is that the behavior to reduce zoom is learned so as to track different kinds of vehicles in the presence of other objects/occlusions and scene variations. Due to the reduction in $Obj_{size}$, the average training reward decreases from Sc-1 to Sc-5.

|  | Fixed background | | | | Variable backgrounds+Trees | | | | Variable backgrounds+Trees+Object | | | |
| --- | --- | --- | --- | --- | --- | --- | --- | --- | --- | --- | --- | --- |
|  | %Tracking | $Center_x$ | $Center_y$ | $Obj_{size}$ | %Tracking | $Center_x$ | $Center_y$ | $Obj_{size}$ | %Tracking | $Center_x$ | $Center_y$ | $Obj_{size}$ |
| $DT_v$ | 98.8 ± 7.6 | 0.86 | 0.85 | 0.30 | 92.0 ± 18.7 | 0.85 | 0.84 | 0.30 | 88.7 ± 21.4 | 0.85 | 0.85 | 0.29 |
| $DT_h$ | 95.8 ± 15.2 | 0.90 | 0.90 | 0.27 | 86.3 ± 22.7 | 0.89 | 0.90 | 0.28 | 83.6 ± 23.7 | 0.89 | 0.90 | 0.28 |

Table 5. The dynamic tasking (DT) performance of Eagle policies to track either a humans character ($DT_h$) or a vehicle ($DT_v$). $Center_x$, $Center_y$ and $Obj_{size}$ have std within ±0.1.

| Height | 20m | 10m | 8m | 5m | 4m |
| --- | --- | --- | --- | --- | --- |
| %Tracking | 96.5 ± 15 | 99.7 ± 3.1 | 99.7 ± 5.4 | 99.3 ± 6.6 | 90.2 ± 23 |

Table 6. %Tracking of Eagle policies at different heights (meters) of the PTZ camera.

**Generalization to unseen scene variations**: Eagle policies for Sc-1 and Sc-2 are trained using a *Fixed background*. We see these policies don't transfer to unseen scenes variations (*Variable backgrounds+Trees* and *Variable backgrounds+Trees+Humans* as shown in Table 3). Adding *Image augmentations* in Sc-2 results in better performance in comparison to Sc-1 for unseen variations. During training, Sc-3 doesn't observe human characters; however, the policy maintains a tracking duration of 90% even in the presence of human characters. This shows variations in backgrounds are critical during training to have generalizability. We also see that Sc-4 and Sc-5 have similar behavior. This is because Sc-4 and Sc-5 are trained with variable backgrounds, trees, and human characters for $SUV1_{blue}$ vehicle. However, Sc-5 is also trained to track rich categories of vehicles, and next, we see how Sc-4 and Sc-5 generalize to unseen vehicle variations.

**Generalization to unseen variations in an object of interest**: We analyze Sc-4 and Sc-5 policies to track vehicles not present during training. The performance is shown in Table 4 to track $HatchBack_{green}$ and $Truck_{blue}$. *HatchBack* and *Truck* vehicles were not present during training. Also, no green color vehicles were present during training. The policies trained for Sc-5 outperform because Sc-5 has different types of vehicles with multiple colors, whereas Sc-4 has one type of vehicle with multiple colors. Hence, having variations in vehicle types generalizes better. Further, $HatchBack_{green}$ refers to a setting where neither the vehicle (*HatchBack*) nor the color (*green*) was present during training, which gives slightly worse performance than $Truck_{blue}$. *blue* color vehicles were present during the training of policies.

**Dynamic Tasking Performance**: Dynamic tasking represents the most complex scenario in Table 2 as the agent needs to learn adaptation of its tracking goal across diverse objects. Table 5 shows the performance of Eagle policies to track either $SUV1_{blue}$ ($DT_v$) or a *Human* ($DT_h$) character. In *Fixed background* and *Variable backgrounds+Trees* only one of the objects of interest ($SUV1_{blue}$ or a *Human*) is present. In *Variable backgrounds+Trees+Object* both $SUV1_{blue}$ and *Humans* are present. In the absence of another competing object, policy performs much better, whereas it suffers a performance degradation in *Variable backgrounds+Trees+Object*. The tracking goal also has different complexity between the vehicle and human sub-class. The vehicle sub-class contains only a single vehicle ($SUV1_{blue}$), whereas the human sub-class contains 4 different characters. The average performance of 4 human characters is shown. We hypothesize that the imbalance in tracking complexity results in performance differences between $DT_v$ and $DT_h$.

**Camera height**: Eagle policies are trained by placing a camera at 8 meters(m) height as discussed in Section 5.3 (Tracking setup). Table 6 shows the performance of Sc-5 policies trained at 8m by varying the height during evaluation.

| Approach | Fixed background | | | | Variable backgrounds+Trees+Humans | | | |
|---|---|---|---|---|---|---|---|---|
| | %Tracking | $Center_x$ | $Center_y$ | $Obj_{size}$ | %Tracking | $Center_x$ | $Center_y$ | $Obj_{size}$ |
| Eagle | 99.7 ± 5.4 | 0.86 ± 0.1 | 0.85 ± 0.1 | 0.27 ± 0.1 | 99.2 ± 4.6 | 0.86 ± 0.1 | 0.85 ± 0.1 | 0.27 ± 0.1 |
| Yolo+Kalman+Controller | 95.1 ± 18.2 | 0.81 ± 0.1 | 0.85 ± 0.1 | 0.24 ± 0.1 | 75.4 ± 30.6 | 0.81 ± 0.1 | 0.84 ± 0.1 | 0.25 ± 0.1 |
| Yolo+Deep_RL | 39.1 ± 18.9 | 0.81 ± 0.2 | 0.82 ± 0.2 | 0.19 ± 0.2 | 36.8 ± 19.0 | 0.84 ± 0.2 | 0.82 ± 0.2 | 0.20 ± 0.2 |
| NN+Controller | 81.2 ± 26.1 | 0.83 ± 0.1 | 0.82 ± 0.1 | 0.22 ± 0.1 | 53.6 ± 31.5 | 0.86 ± 0.1 | 0.85 ± 0.1 | 0.23 ± 0.1 |

Table 7. Comparison of Eagle with the current state-of-the-art approaches for different scene complexities.

| Approach | %Tracking | $Center_x$ | $Center_y$ | $Obj_{size}$ |
|---|---|---|---|---|
| PerfectBB+Deep_RL | 98.5 ± 8.5 | 0.85 ± 0.1 | 0.81 ± 0.2 | 0.36 ± 0.1 |
| PerfectBB+Kalman+Controller | 98.3 ± 8.4 | 0.85 ± 0.1 | 0.83 ± 0.1 | 0.33 ± 0.1 |

Table 8. Performance of multi-stage approaches when perfect bounding boxes are available from EagleSim simulator.

The %Tracking is captured across 100 episodes to track $SUV1_{blue}$ vehicle moving on random trajectories (avg speed of 6 m/s) in scenes having *Fixed Background*. The action space of Eagle modifies the current PTZ parameters (Section 5.2) and works well between 5m to 20m heights during deployment, even when trained at 8m. We do observe performance drop as the camera placement differs from the training setup. This suggests with a substantial camera placement difference between training and deployment; policy retraining is needed for optimal performance, which is similar to the recalibration of PTZ parameters in classical control.

**Object of interest in the initial viewpoint**: Eagle policies successfully track the object of interest even when the initial viewpoint doesn't capture the object of interest completely or has another object which is captured in dominance. Figure 5 shows a scene where the blue car is captured in dominant in the initial viewpoint and is partially blocking the object of interest (red car). Both cars are exactly similar except for their colors. As seen, in the subsequent images, Eagle successfully tracks the red car.

**Summary**: *What is the impact on the PTZ controller performance as the tracking complexity is increased?* As the PTZ tracking complexity is increased for vehicle tracking scenarios (Sc-1 to Sc-5), Eagle policies learn a more conservative behavior. The zoom level maintained on the object of interest is reduced in complex scenes. Policies trained in simpler scenes don't transfer well to unseen backgrounds and cannot work in the presence of other objects like trees or humans. The trained policies for Sc-5 use complex scenes, which also perform better for unseen variations in the object of interest. This shows the complex scenes in EagleSim simulator are critical to developing generalizable policies. The evaluation of dynamic tasking showed that contextual input enables goal modification during the deployment; however, enabling dynamic tasking comes at a performance cost.

### 5.4 Eagle vs Other Approaches

Here, we compare the performance of Eagle with the current state-of-the-art approaches. First, we discuss our realization of other approaches and then present their control performance.

*5.4.1 Object_detection+tracking+control.* We use Yolo5s [2], a state-of-the-art lightweight object detector, followed by a Kalman filter for state estimation on bounding boxes. We use the open-source [6] implementation of the SORT algorithm for the Kalman filter. Finally, the tracking outputs are used to control the PTZ parameters using a rule-based controller [15]. The input to the object detector is an image of size 240×240. We call this tracking setting *Yolo+Kalman+Controller*.

Yolo5s is trained to detect 80 different classes of objects [2]. To improve its performance during evaluation, we finetune Yolo5s by collecting a labeled dataset of 50k images from EagleSim. The dataset is collected by observing the $SUV1_{blue}$ vehicle in *Variable backgrounds+Trees+Humans* scenes and at varying zoom levels. The default hyperparameters recommended by Yolo5s developers [2] were used to finetune the model for 100 epochs. We use the finetuned Yolo5s in our experiments.

Finally, to completely remove object detector errors, we implement another setting that uses the oracle bounding boxes provided by the EagleSim simulator as input to the Kalman filter. We call this setting, *PerfectBB+Kalman+Controller*, where PerfectBB signifies the perfect bounding boxes.

**Controller Parameter Tuning**: We fine-tune the parameters of this pipeline using Mango [22], a state-of-the-art hyperparameter tuning library. We use total reward in Equation 4 as the objective function for Mango to tune controllers for *PerfectBB+Kalman+Controller* and *Yolo+Kalman+Controller* for 1000 episodes (each of 2000 steps) in EagleSim. This amounts to a total tuning time of 16.6 hours.

*5.4.2 Object_detection+reinforcement learning.* We use the bounding boxes from EagleSim to train a deep-RL model. The input state of deep-RL consists of a vector of 4 variables ($[Xmin, Ymin, Xmax, Ymax]$) shown in Figure 3. The policy network is a 2-layer fully connected neural network, each having 64 hidden nodes, followed by an output layer with 3 discrete outputs for pan, tilt, and zoom parameters. PPO algorithm with the same hyperparameters as Eagle is used for training (Section 5.2).

We evaluate this approach at test time in two different ways: (i) using the bounding boxes from EagleSim, which are error-free, called *PerfectBB+Deep_RL*, and (ii) using the bounding boxes from the finetuned Yolo5s object detector, *Yolo+Deep_RL*. The input to object detector is an image of size 240×240. This shows the usage of object detectors for a realistic setting when perfect bounding boxes are not available.

*5.4.3 Relative_location+control.* We build on the method proposed by Kyrkou et al. [14] using supervised machine learning for pan-tilt control. We extend the proposed [14] approach by including a relative zoom variable. A neural network is trained to predict three outputs defining a relative location: (i) $Rel_X = \frac{x}{Width/2}$, (ii) $Rel_Y = \frac{y}{Height/2}$, and (iii) $Rel_{Zoom} = \frac{(Xmax-Xmin)*(Ymax-Ymin)}{Width*Height}$. Where $x, y, Width, Height, Xmax, Xmin, Ymax$ and $Ymin$ are shown in Figure 3. The $Rel_X$, $Rel_Y$, and $Rel_{Zoom}$ varies between -1 and 1. A separate controller uses the $Rel_X$ and $Rel_Y$ to control the pan and tilt of the camera and uses $Rel_{Zoom}$ to modify the zoom.

The neural network has the same architecture as the policy network of Eagle (Table 1), with a different output layer. The output layer consists of 3 nodes for each of $Rel_X$, $Rel_Y$, and $Rel_{Zoom}$ with a linear activation. The input to the network is an image of size 240×240. This setting is called *NN+Controller*. We train this neural network by using the 50k labeled images capturing the relative location of $SUV1_{blue}$ vehicle in *Variable backgrounds+Trees+Humans* scenes from the EagleSim simulator.

*5.4.4 Performance Comparisons.* The comparison of Eagle with the current state-of-the-art approaches is shown in Table 7. The performance numbers for Eagle are added for policies trained for the scenario Sc-5 (from Figure 3). For fairness, all approaches are evaluated for an end-to-end delay of 30ms from sensing image to applying PTZ actions. The evaluation in Table 7 is to track $SUV1_{blue}$ vehicle moving on random trajectories (with an average speed of 6 m/s) for 100 episodes The tracking setup used for evaluation is the same as discussed in Section 5.3 (Tracking setup).

The performance of multi-stage approaches using perfect bounding boxes is shown in Table 8. The accuracy of perfect bounding boxes is independent of the scene's complexity. But, these settings are not realistic, as, in practice, there will be errors in object detectors; however, this evaluation presents a valuable insight into the upper bound of performance as the object detectors improve.

In *Fixed background*, Eagle outperforms the next best approach of *Yolo+Kalman+Controller* by 4.6% in tracking duration (% Tracking). Eagle also maintains the metrics of $Center_x$, $Center_y$, and $Obj_{size}$ better than others. With perfect bounding boxes (Table 8), controller performance improves as expected. However, when bounding boxes are imperfect (Table 7), there is a significant degradation. *Yolo+Deep_RL* suffers more degradation in comparison to *Yolo+Kalman+Controller*. This is because, for *Yolo+Kalman+Controller*, Mango (hyperparameter tuner) selects lower $Obj_{size}$ during tuning to maintain a higher tracking duration. However, in *Yolo+Deep_RL*, such fine-tuning is not possible, and on replacing perfect bounding boxes with Yolo5s. We see Eagle slightly outperforms even the *PerfectBB+Kalman+Controller* and *PerfectBB+Deep_RL* in tracking duration; we hypothesize that the raw images provide much richer information such as the orientation of the vehicle and the presence of other objects, which is not available in perfect bounding boxes.

In complex scenes (*Variable backgrounds+Trees+Humans*), Eagle outperforms the following best approach (*Yolo+Kalman+Controller*) by 23.8% in tracking duration (% Tracking) and also maintains other metrics superior. This is because, in more complex scenes, object detectors face more challenges in identifying the object of interest.

| Approach | Variable backgrounds+Trees+Humans | | | |
|---|---|---|---|---|
| | %Tracking | $Center_x$ | $Center_y$ | $Obj_{size}$ |
| Eagle | 99.2 ± 4.6 | 0.86 | 0.85 | 0.27 |
| *Object Detector(150k)* | 82.0 ± 29.6 | 0.85 | 0.84 | 0.28 |
| *Object Detector(150k)* | 81.8 ± 28.1 | 0.85 | 0.84 | 0.27 |
| *Object Detector(50k)* | 78.5 ± 31.2 | 0.84 | 0.85 | 0.27 |

Table 9. Comparison of Eagle with PTZ trackers using custom object detectors.

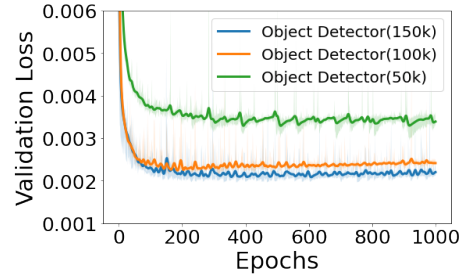

Fig. 6. Learning curves of custom object detectors.

### 5.5 Custom Lightweight Object Detectors

The traditional multi-stage pipelines adopt complex neural object detectors such as YOLO, ResNet-based, and faster R-CNN based as discussed in Section 2.1. Here, we compare Eagle with pipelines using very lightweight custom object detectors trained from scratch for a specific deployment. Training custom object detectors from scratch demands costly effort to label bounding boxes; hence, it is challenging to realize in practice. However, the introduced EagleSim simulator allows us to train custom object detectors by providing ground-truth annotations.

We develop custom object detectors with network architecture similar to Eagle (Table 1) with a modified output layer. The output layer directly predicts bounding boxes (Xmin, Ymin, Xmax, Ymax) of vehicles in the images. We generate labeled bounding box image datasets of sizes 50k, 100k, and 150k using EagleSim by driving 6 different vehicles in the Sc-5 scenario (Table 2) using the tracking setup discussed in Section 5.3.

**Learning curves**: We train the networks using the mse loss function and adam optimizer for 1000 epochs (Figure 6). Three checkpoints for each network with different random seeds are trained using 20% of the data as the validation. In Figure 6, *Object Detector(150k)* refers to the model trained using the 150k image dataset.

**PTZ tracking performance**: A PTZ tracker is realized by using lightweight custom object detectors followed by a Kalman filter and a rule-based controller, as discussed in Section 5.4.1. Table 9 compares the performance of Eagle with these custom multi-stage pipelines. We see that Eagle outperforms the specialized lightweight detectors significantly by maintaining up to 17% more tracking duration. We hypothesize that detecting accurate bounding boxes becomes more challenging in the presence of other objects in the scenes. We evaluate different training dataset sizes and observe a slight improvement in tracking performance as the dataset size is increased.

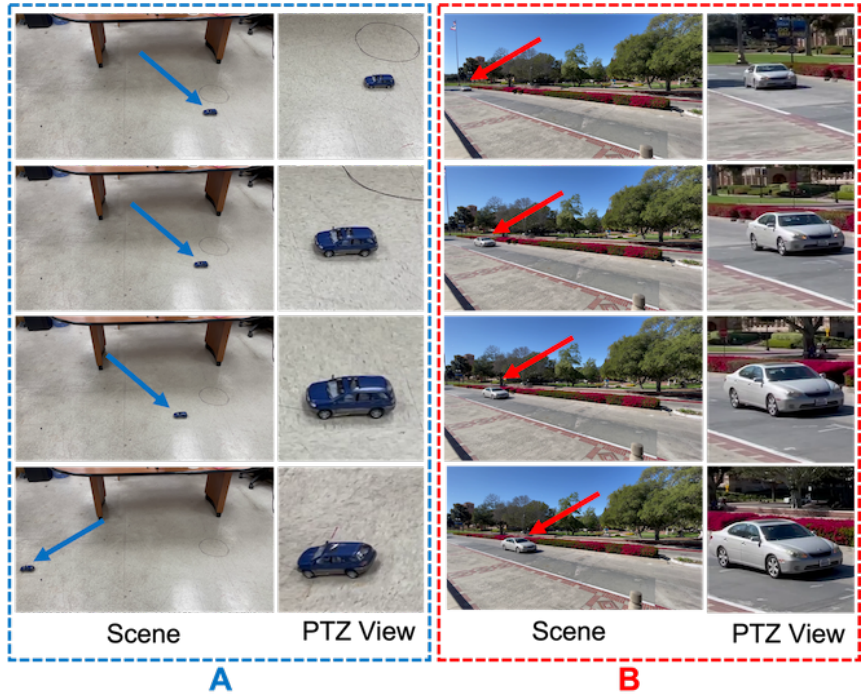

Fig. 7. Eagle policies on real videos. The arrows show the vehicle to track in the video scene. The PTZ view is maintained by Eagle while tracking the vehicle. The top images show the starting point where the PTZ view is not focused.

### 5.6 Transfer of Eagle to the Real Scene Videos

We show the direct transfer of Eagle policies from simulation to the real scene videos. EagleSim support capability to simulate pan-tilt-zoom actions on real videos. This is realized by simulating the effect of policy actions on an initial wider view containing an object of interest in the video. This evaluation presents a visual way to see the behavior of Eagle policies on real scenes.

We simulate the pan-tilt-zoom actions from Eagle policies trained for Sc-5 on real-scene videos. Tracking videos for real scenes are available online[2]. The results are shown in Figure 7 for two different scenes. The left scene consists of a toy blue SUV vehicle moving on a concrete floor. The variations in this scene are much simpler. The tracking progress is shown from the top image to the bottom image in Figure 7:A. The initial bounding box of wide view (PTZ View) is manually selected, which is given as input to the trained policy. The actions predicted by the policy are used to update the selected bounding box by moving it left-right for pan and up-down for tilt. The zoom action controls the size of the bounding box. The scene in Figure 7:B shows a real vehicle of grey color moving in a background having trees/patterns. The tracking progress is again shown from top to bottom, where the first image of PTZ View shows the wider view given as an input to the policy. As the tracking progresses, the policy follows the object of interest (car). For optimal real-world performance, policy training can be complemented using real-world finetuning. However, here we focus on a simulation environment and show out-of-box transfer of Eagle policies to the real scene videos.

| Device | Raspberry Pi 4B | Jetson Nano |
|--------|-----------------|-------------|
| Eagle  | 9.2 ± 1.3 ms    | 6.2 ± 2 ms  |
| Yolo5s | 1817 ± 14.1 ms  | 86.2 ± 1.5 ms |

Table 10. Inference time in milliseconds (ms) of Eagle and optimized Yolo5s on embedded camera platforms.

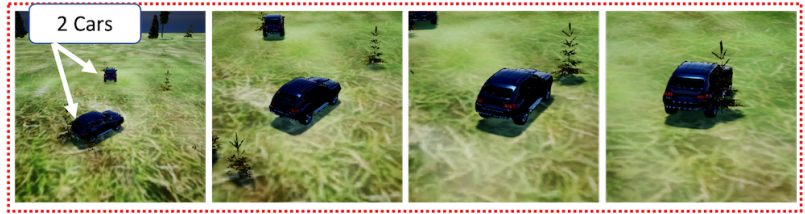

Fig. 8. A sample scene with multiple objects of interest.

### 5.7 Runtime of Eagle on Embedded Cameras

We measure the runtime of Eagle policies on Raspberry Pi 4B and Jetson Nano devices which are also supported by embedded PTZ cameras [1]. We compare Eagle with the next best traditional approach using a general-purpose object detector from Table 7: *Yolo+Kalman+Controller* . Eagle policies have only 79k network parameters compared to Yolo5s' 7.2 million parameters. The inference latency of Eagle and Yolo5s is reported in Table 10. On Raspberry Pi, we optimize the inference for Eagle and Yolo5s using TensorFlow Lite[6]. On Jetson Nano, the Yolo5s model has an inference latency of 218 milliseconds (ms). We optimize Yolo5s for Jetson Nano using TensorRT[7] and further, quantize the model to float16 to reduce its inference latency. The inference latency of optimized Yolo5s on Jetson Nano is reported in Table 10.

The PTZ camera [1] supports a frame rate of 120 Hz, which when tested with neural network inference reduces to 100 Hz. The camera supports PTZ actions with a resolution of 1 degree and has an actuation delay of 10 ms. When using Eagle for PTZ control on Raspberry PI, the end-to-end delay is around 30 ms (10 ms for inference, 10 ms for sensing (100 Hz), and 10 ms for actuation), enabling a real-time deployment with 33 FPS. Yolo5s has an inference latency of 1817 ms on Raspberry PI, making it completely infeasible to run *Yolo+Kalman+Controller* on Raspberry PI.

On Jetson Nano, Eagle achieves an even higher FPS of 38 (inference latency of 6.2ms and similar sensing+actuation delay of 20ms). *Yolo+Kalman+Controller*, even when using optimized Yolo5s, has a significantly higher inference latency of 86.2ms on Jetson Nano. Considering the sensing and actuation delays, the total end-to-end delay for *Yolo+Kalman+Controller* is 106.2 ms, which results in a very low FPS of 9. Hence, for real-time deployment, Eagle policies represent a lightweight controller with superior control performance.

## 6 DISCUSSION

**Policy behavior on multiple objects of interest**: Eagle policies in Section 3.3 are trained by assuming a single object of interest from class *A* is present in the scene, while the same policy generalizes across all objects of class *A*. Here, we test the trained policy by adding multiple objects from class $A$ to the same scene, as shown in Figure 8. We see that the reward of Equation 1 incentivizes an agent to track the larger object lying closer to the center of the image to achieve a higher expected sum of rewards. We test the policy of the Sc-5 scenario. As seen in the images from left to the right, the policy tracks the larger car and continues tracking it after it gets focused.

---
[6]https://www.tensorflow.org/lite
[7]https://github.com/tensorflow/tensorrt

**Speed of objects**: We evaluated controller performance on vehicles moving at an average speed of 6m/s (max speed of 16m/s). The action space of Section 5.2 also tracks slow-moving humans. To track even faster-moving objects, the action space can have more options, or the end-to-end delay can be reduced to modify the PTZ parameters faster.

**Multiple cameras**: EagleSim can capture bounding boxes of multiple objects of interest and also allows the control of multiple PTZ cameras in the same scenes. Multiple cameras present a problem of collaborative tracking. We leave the study of end-to-end controllers for collaborative tracking as future exploration.

**Limitations of Eagle**: The multi-stage pipeline is modular, allowing the replacement of each stage with several alternatives; however, the end-to-end controller cannot be modified in this fashion. Changing the tracking goal in a multi-stage pipeline is easier by filtering the outputs from object detectors. Modifying the tracking goal in Eagle to a new type of object may require retraining from scratch. EagleSim simulator is designed to automate the retraining and alleviate the labeling efforts. We also saw that runtime tasking in Eagle to selectively track humans and vehicles comes at a performance cost. Runtime tasking in multi-stage approaches can be enabled by changing their tracking goal by adopting general-purpose object detectors. In light of these limitations, Eagle is more suitable for applications where the general categories of objects to track are known ahead, and lightweight controllers are required.

# 7 CONCLUSION

We introduced a new lightweight approach called Eagle that outperforms prior state-of-the-art and provides superior end-to-end PTZ control performance. We benchmark Eagle on different scenarios with varying tracking complexity and evaluate its generalizability on unseen objects/surroundings. To automate the training and evaluation of PTZ camera controllers, we also presented an accompanying simulator called EagleSim. Further, the availability of Oracle bounding boxes in EagleSim enables us to study prior multi-stage approaches by removing errors in their object detection pipeline showing upper-performance bounds. End-to-end control enabled by Eagle can slightly outperform these upper bounds. Finally, the action predicted by end-to-end Eagle policies trained purely in photo-realistic simulation can transfer to real-world videos, suggesting it is a promising alternative to the current approaches.

**ACKNOWLEDGMENTS**

The research reported in this paper was sponsored in part by the Air Force Office of Scientific Research (AFOSR) under Cooperative Agreement FA9550-22-1-0193; the IoBT REIGN Collaborative Research Alliance funded by the Army Research Laboratory (ARL) under Cooperative Agreement W911NF1720196; the National Science Foundation (NSF) under awards # 2211301 and # 2124252; and, the CONIX Research Center, one of six centers in JUMP, a Semiconductor Research Corporation (SRC) program sponsored by DARPA. The views and conclusions contained in this document are those of the authors and should not be interpreted as representing the official policies, either expressed or implied, of the AFOSR, ARL, DARPA, NSF, SRC, or the U.S. Government. The U.S. Government is authorized to reproduce and distribute reprints for Government purposes notwithstanding any copyright notation here on.